\documentclass{tlp}
\usepackage{tikz}
\usetikzlibrary{positioning, arrows.meta, shapes.geometric}

\usepackage{amsmath}
\usepackage{graphicx}
\usepackage{multirow}
\usepackage{booktabs}
\usepackage{amsfonts}
\begin{document}

\lefttitle{Cambridge Author}

\jnlPage{1}{8}
\jnlDoiYr{2021}
\doival{10.1017/xxxxx}

\title[Symbolic Rule Extraction from Attention-Guided Sparse Representations in ViTs]{Symbolic Rule Extraction from Attention-Guided Sparse Representations in Vision Transformers\thanks{This work was supported by US NSF Grant IIS 1910131, US DoD and industry grants. We thank the members of the ALPS lab at UT Dallas for insightful discussions.}}

\begin{authgrp}
\author{\sn{Parth} \gn{Padalkar}}
\affiliation{The University of Texas at Dallas}
\author{\sn{Gopal} \gn{Gupta}}
\affiliation{The University of Texas at Dallas}
\end{authgrp}

% \history{\sub{xx xx xxxx;} \rev{xx xx xxxx;} \acc{xx xx xxxx}}

\maketitle

\begin{abstract}
Recent neuro-symbolic approaches have successfully extracted symbolic rule-sets from CNN-based models to enhance interpretability. However, applying similar techniques to Vision Transformers (ViTs) remains challenging due to their lack of modular concept detectors and reliance on global self-attention mechanisms. We propose a framework for symbolic rule extraction from ViTs by introducing a sparse concept layer inspired by Sparse Autoencoders (SAEs). This linear layer operates on attention-weighted patch representations and learns a disentangled, binarized representation in which individual neurons activate for high-level visual concepts. To encourage interpretability, we apply a combination of L1 sparsity, entropy minimization, and supervised contrastive loss. These binarized concept activations are used as input to the FOLD-SE-M algorithm, which generates a rule-set in the form of logic programs. Our method achieves a $\textbf{5.14\%}$ better classification accuracy than the standard ViT while enabling symbolic reasoning. Crucially, the extracted rule-set is not merely post-hoc but acts as a logic-based decision layer that operates directly on the sparse concept representations. The resulting programs are concise and semantically meaningful. This work is the first to extract executable logic programs from ViTs using sparse symbolic representations. It bridges the gap between transformer-based vision models and symbolic logic programming, providing a step forward in interpretable and verifiable neuro-symbolic AI.
\end{abstract}


\section{Introduction}
Extracting logic-based rules from neural models has emerged as a central objective in neuro-symbolic AI, driven by the growing demand for interpretability and verifiability in machine learning. As deep learning models continue to scale, they are increasingly deployed in critical applications such as autonomous driving (\cite{kanagaraj2021deep}), medical diagnosis (\cite{Sun2016ComputerAL}), and natural disaster prevention (\cite{Ko2012disaster}). In these sensitive domains, incorrect predictions can carry severe consequences, emphasizing the necessity for transparency in decision-making. Many of these applications depend significantly on accurate image classification models, particularly Convolutional Neural Networks (CNNs). Recent frameworks such as NeSyFOLD (\cite{nesyfold,nesyfold-g}) and ERIC (\cite{eric}) have successfully demonstrated the extraction of human-interpretable symbolic rule sets from CNNs, providing insights into the underlying reasoning of predictions in vision tasks.

Most of this progress, however, has been concentrated on CNNs. CNNs are composed of filters, that are trainable matrices that learn to detect patterns in local regions of images (e.g., a bed in an image classified as a bedroom). Their modular architecture, where individual filters often correspond to distinct visual concepts, makes them particularly suitable for rule extraction. By binarizing the activations of the final layer and using them as input to rule-based machine learning algorithms such as decision trees or FOLD-SE-M (\cite{foldsem}), it is possible to extract the symbolic rule-sets.

In contrast, Vision Transformers (ViTs) (\cite{vit}) have remained largely inaccessible to symbolic extraction techniques. While ViTs now dominate the field of vision due to their superior performance and flexibility, their reliance on global self-attention and lack of explicit concept detectors pose a significant challenge for rule-based interpretability. ViTs encode information in a distributed and entangled manner, making it unclear how to localize or discretize concept-level representations.

In this work, we take a step toward bridging this gap by introducing a framework—NeSyViT—for extracting logic-based rule sets from Vision Transformers (ViTs). We modify the standard ViT architecture by replacing the final fully connected classification head with a single linear layer--\textit{sparse concept layer}--trained to produce binarized outputs. The goal is to encourage each neuron in this final layer to correspond to a few distinct high-level concepts. To achieve this, we draw inspiration from Sparse Autoencoders (SAEs) (\cite{sae}), incorporating an L1 sparsity loss to ensure that only a small subset of neurons activates for any given image.

To encourage binarization, we apply a sigmoid activation to the linear outputs, constraining them to the $[0, 1]$ range, and introduce an entropy-based loss that pushes activations toward binary extremes (0 or 1). This design enables us, after training, to extract a binary vector for each image that reflects concept-level activations. These vectors are then passed to the FOLD-SE-M algorithm to generate a symbolic rule-set in the form of a stratified Answer Set Program for classification. It is crucial that the representations corresponding to images from the same class are well-clustered in the latent space. This promotes the formation of clear decision boundaries that can be effectively exploited by FOLD-SE-M. To enforce this structure, we incorporate a supervised contrastive loss (SupCon) (\cite{supcon}), which encourages representations of images with the same label to lie closer together while pushing apart those from different classes. This facilitates learning highly accurate rules. 

\begin{figure}
    \centering
    \includegraphics[width=\linewidth]{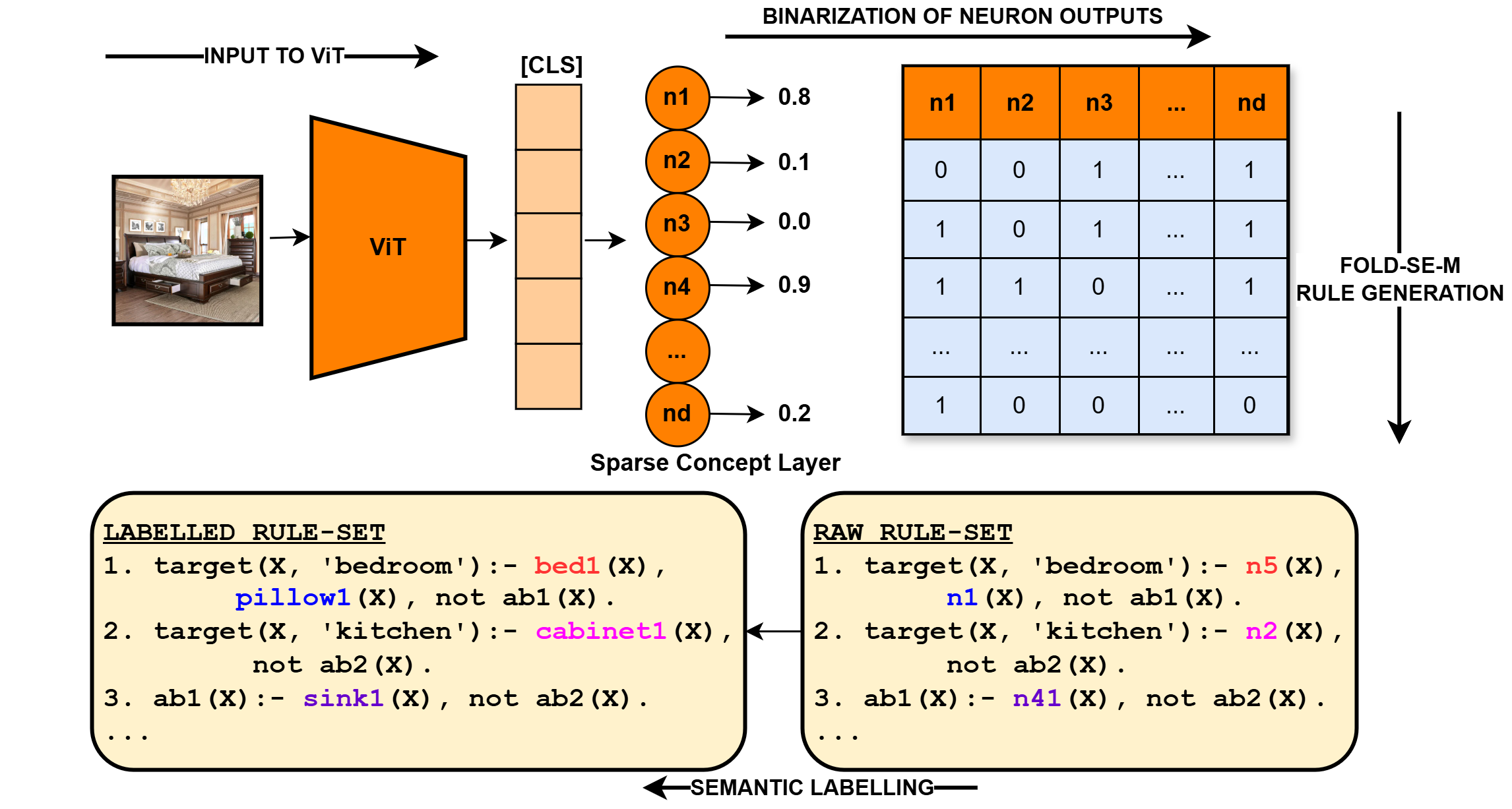}
    \caption{The NeSyViT Framework}
    \label{fig_nesyvit}
\end{figure}

Finally, each predicate in the extracted rule set is ``semantically labelled" based on the concept(s) that its corresponding neuron represents. Padalkar et al. introduced an algorithm for automatic semantic labelling of the predicates in the rule-sets extracted from a CNN using FOLD-SE-M (\cite{nesyfold}). We adopt the same algorithm for semantic labelling of the rule-sets we generate. Figure \ref{fig_nesyvit} illustrates the NeSyViT framework.

% Unlike prior work such as NeSyFOLD, which binarizes neuron activations post hoc—often losing information when removing deeper layers—our method learns binarized representations during training, preserving semantic content and enabling more faithful symbolic reconstruction.
The final \underline{Ne}uro-\underline{Sy}mbolic (NeSy) model is a combination of the ViT based feature extractor that generates the binary vector for each image and the rule-set generated by FOLD-SE-M, which classifies the image into a particular class based on which neurons were activated/deactivated (1/0). We demonstrate through our experiments that the NeSy model produced by our framework outperforms the vanilla ViT in terms of classification accuracy, achieving an average improvement of $\textbf{5.14\%}$. Notably, this performance gain is achieved while also generating interpretable rule-sets—a significant result, as prior neuro-symbolic frameworks that extract rule-sets from neural models often suffer a drop in accuracy.

Our contributions are as follows:
\begin{enumerate}
    \item We introduce a training method that combines supervised contrastive loss, L1 sparsity, and entropy loss to learn compact and binarized concept-level representations.
    \item We propose an end-to-end neuro-symbolic framework, \textit{NeSyViT}, for generating a rule-set from a modified ViT using FOLD-SE-M.
    \item We show through experiments that our NeSyViT framework achieves classification accuracy better than the vanilla ViT, while also producing concise rule-sets
\end{enumerate}

\section{Background}
\subsection{Vision Transformers}

The Vision Transformer (ViT) (\cite{vit}) is a deep learning architecture that applies the Transformer model—originally developed for natural language processing—to image classification tasks. Unlike Convolutional Neural Networks (CNNs), which operate on local regions using convolutional filters, ViTs treat an image as a sequence of patches. An input image is divided into fixed-size non-overlapping patches (e.g., $16 \times 16$ pixels), each of which is flattened and projected into a vector embedding. The vector embeddings of those patches then go through multiple \textit{self-attention} layers followed by the final classification layer. 

The Transformer uses self-attention layers to model relationships between all image patches simultaneously. At the core of this mechanism is the concept of an \textit{attention head}, which determines how much attention each patch should pay to every other patch in the sequence. For each patch token, the model learns three distinct vectors: a \textit{query}, a \textit{key}, and a \textit{value}. Attention scores are computed by taking the dot product between a patch’s query vector and the key vectors of all other patches, followed by a softmax normalization to produce attention weights. These weights are then used to compute a weighted sum of the value vectors, generating a context-aware representation for each token.

A special learnable token, known as the \texttt{[CLS]} token, is prepended to the patch sequence before being passed into the Transformer. Unlike the patch tokens that represent image regions, the \texttt{[CLS]} token acts as a summary placeholder. During self-attention, it aggregates information from all other patches by attending to them across multiple layers. After the final Transformer layer, the embedding corresponding to the \texttt{[CLS]} token is typically used for classification, as it encapsulates a global representation of the image.

While this architecture allows Vision Transformers to capture long-range dependencies and global structure, it also presents a major challenge for interpretability. In CNNs, individual filters often learn to detect localized visual patterns or objects (e.g., corners, textures, faces), enabling a degree of modularity and conceptual traceability. In contrast, attention heads are distributed, context-sensitive, and dynamically influenced by the full set of tokens—including the \texttt{[CLS]} token. This makes it difficult to associate specific heads or tokens with interpretable visual concepts. Consequently, extracting symbolic, modular representations from ViTs is significantly more challenging than from CNNs.
Hence, in this work, we take the \texttt{[CLS]} vector and pass it through the sparse concept layer that is optimized to produce sparse outputs, allowing the information encoded in the \texttt{[CLS]} token to be disentangled into a small set of active neurons. Each neuron is encouraged to specialize and activate primarily for images of a specific class. This design is inspired by the core principles of Sparse Autoencoders (SAEs) (\cite{sae}), which promote compact and interpretable representations through sparsity.

\subsection{FOLD-SE-M}
The FOLD-SE-M algorithm \cite{foldsem} that we employ in our framework, learns a rule-set from data as a \textit{default theory}.
Default logic is a non-monotonic logic used to formalize commonsense reasoning. A default $D$ is expressed as:
  
\begin{equation}\label{eq_1}
    D = \frac{A: \textbf{M} B}{\Gamma}
\end{equation}
 
\noindent Equation \ref{eq_1} states that the conclusion $\Gamma$ can be inferred if pre-requisite $A$ holds and $B$ is justified. $\textbf{M} B$ stands for ``it is consistent to believe $B$".
Normal logic programs can encode a default theory quite elegantly (\cite{gelfondkahl}). A default of the form: 
$$\frac{\alpha_1 \land \alpha_2\land\dots\land\alpha_n: \textbf{M} \lnot \beta_1, \textbf{M} \lnot\beta_2\dots\textbf{M}\lnot\beta_m}{\gamma}$$
\noindent can be formalized as the
normal logic programming rule:
$$\gamma ~\texttt{:-}~ \alpha_1, \alpha_2, \dots, \alpha_n, \texttt{not}~ \beta_1, \texttt{not}~ \beta_2, \dots, \texttt{not}~ \beta_m.$$
\noindent where $\alpha$'s and $\beta$'s are positive predicates and \texttt{not} represents negation-as-failure. We call such rules \emph{default rules}. 
Thus, the default 

$$\frac{bird(X): M \lnot penguin(X)}{flies(X)}$$

\noindent will be represented as the following default rule in normal logic programming:

{\tt flies(X) :- bird(X), not penguin(X).}

\noindent We call {\tt bird(X)}, the condition that allows us to jump to the default conclusion that {\tt X} flies, the {\it default part} of the rule, and {\tt not penguin(X)} the \textit{exception part} of the rule. 

FOLD-SE-M (\cite{foldsem}) is a Rule Based Machine Learning (RBML) algorithm. It generates a rule-set from tabular data, comprising rules in the form described above. The complete rule-set can be viewed as a stratified answer set program (a stratified ASP rule-set has no cycles through negation (\cite{Baral})). It uses special {\tt abx} predicates to represent the exception part of a rule where {\tt x} is a unique numerical identifier.   
FOLD-SE-M incrementally generates literals for \textit{default rules} that cover positive examples while avoiding covering negative examples. It then swaps the positive and negative examples and calls itself recursively to learn exceptions to the default when there are still negative examples falsely covered.

There are $2$ tunable hyperparameters, $ratio$, and $tail$. 
The $ratio$ controls the upper bound on the number of false positives to the number of true positives implied by the default part of a rule. The $tail$ controls the limit of the minimum number of training examples a rule can cover.

% \begin{figure}[t]
%     \centering
%     \includegraphics[width=0.9\linewidth]{figs/vit_diagram.pdf}
%     \caption{Overview of the Vision Transformer (ViT) architecture. The image is split into patches and embedded before being passed through multiple layers of self-attention. The output of the classification token is used for final prediction.}
%     \label{fig:vit}
% \end{figure}

\section{Methodology}
Our aim is to extract a symbolic rule-set from the ViT and use it for final classification. To this end, we first modify the architecture of a standard Vision Transformer by replacing the final classification head—placed after the last self-attention layer—with a single linear layer of dimension $D$. The \texttt{[CLS]} token's vector which has the aggregated global information feeds directly into this layer. This layer outputs a $D$-dimensional vector, to which we apply a sigmoid function element-wise, constraining all values to lie in the range $[0,1]$. Each value in this vector could then be interpreted as the activation of a distinct neuron.

There are three key objectives in designing this representation:
\begin{enumerate}
    \item The model needs to produce similar binary vectors for images of the same class, and dissimilar ones for different classes, to allow the FOLD-SE-M to identify meaningful decision boundaries.
    \item The output values had to be pushed as close as possible to either $0$ or $1$, since after training, we binarize these vectors by rounding. Values near $0.5$ would introduce ambiguity and loss of information.
    \item The output vectors need to be sparse, with most values being $0$ and only a few being $1$. This sparsity ensures that rule-sets remain compact and interpretable, as fewer neurons would appear as predicates in the learned rules.
\end{enumerate}

We enforce these three properties using a combination of supervised contrastive loss, entropy minimization, and L1 sparsity loss, described in detail below.

\medskip\noindent\textbf{Supervised Contrastive Loss:}
To ensure that images from the same class produce similar sparse representations, we incorporate the \textit{Supervised Contrastive Loss} (SupCon)~(\cite{supcon}). This loss encourages the $D$-dimensional vectors corresponding to samples of the same class to cluster together in the latent space, while pushing apart those from different classes. Such class-wise clustering is essential for enabling a rule-based learner, such as FOLD-SE-M, to identify clean symbolic decision boundaries.

Let $\mathbf{z}_i \in \mathbb{R}^D$ denote the normalized sparse representation of sample $i$ in a batch $\mathcal{B}$, and let $y_i$ be its corresponding class label. In contrastive learning, we refer to $\mathbf{z}_i$ as the \textit{anchor}, and we compare it to the remaining samples in the batch. Those with the same label are treated as \textit{positives}, while the rest are considered \textit{negatives}.

The SupCon loss for a single anchor $i$ is defined as:

\begin{equation} 
\mathcal{L}_{\text{supcon}}^i = - \frac{1}{|P(i)|} \sum_{p \in P(i)} \log \frac{\exp(\mathbf{z}_i \cdot \mathbf{z}_p / \tau)}{\sum_{a \in \mathcal{A}(i)} \exp(\mathbf{z}_i \cdot \mathbf{z}_a / \tau)}
\end{equation}

where: \begin{itemize} \item $P(i)$ is the set of indices in the batch with the same label as $i$ (i.e., the positives for anchor $i$), \item $\mathcal{A}(i)$ is the set of all indices in the batch excluding $i$ itself, \item $\tau > 0$ is a temperature parameter that controls the sharpness of the distribution. \end{itemize}

The total loss is averaged over all anchors in the batch:

\begin{equation} \mathcal{L}_{\text{supcon}} = \frac{1}{N} \sum_{i \in \mathcal{B}} \mathcal{L}_{\text{supcon}}^i \end{equation}

\noindent where $N$ is the number of images in the batch $\mathcal{B}$.
By minimizing this loss, the model is encouraged to produce sparse vectors that are tightly clustered for images of the same class and distinct from those of other classes. This structure makes the representations well-suited for interpretable rule extraction.

\medskip\noindent\textbf{Entropy Minimization: }
To encourage the sparse representations to become binarized—i.e., close to either 0 or 1—we apply an \textit{entropy minimization} loss on the output of the sigmoid-activated linear layer. Since this linear layer uses a sigmoid function, each neuron's activation lies in the range $[0, 1]$. Ideally, we want these activations to converge toward discrete values ($0$ or $1$) to allow lossless binarization post-training.

We treat each neuron's output as a Bernoulli random variable and compute its entropy using the standard binary entropy formula. Given a batch of $N$ images where $\mathbf{z}_i \in \mathbb{R}^D$ is the $D$-dimensional output for image $i$, the entropy loss is computed as:

\begin{equation} \mathcal{L}_{\text{entropy}} = -\frac{1}{N D} \sum_{i=1}^{N} \sum_{j=1}^{D} \left[ z_{i,j} \log(z_{i,j} + \epsilon) + (1 - z_{i,j}) \log(1 - z_{i,j} + \epsilon) \right], \end{equation}

\noindent where $z_{i,j}$ is the activation of neuron $j$ for image $i$, and $\epsilon$ is a small constant added for numerical stability.

Minimizing this entropy term encourages each activation to move closer to either $0$ or $1$, making the final binarization step more reliable. This is especially important for downstream symbolic rule extraction, where crisp binary features are essential for learning accurate and interpretable logic programs.

\medskip\noindent\textbf{L1 Sparsity Loss: }
To promote interpretability and ensure that only a few neurons activate for each image, we incorporate an \textit{L1 sparsity loss} on the output of the final linear layer. Sparse representations are crucial for generating concise rule-sets, as they limit the number of active predicates per image, reducing the complexity of the learned logic programs.

Formally, given a batch of $N$ vectors ${\mathbf{z}_1, \dots, \mathbf{z}_N}$, where $\mathbf{z}_i \in \mathbb{R}^D$ is the sigmoid-activated output for image $i$, the L1 sparsity loss is defined as:

\begin{equation} \mathcal{L}_{\text{sparsity}} = \frac{1}{N D} \sum_{i=1}^{N} \sum_{j=1}^{D} |z_{i,j}| \end{equation}

This loss penalizes large activation values and encourages most neurons to remain close to zero, allowing only a few dimensions to be active for any given input.

The total loss then becomes:
\begin{equation}
    \mathcal{L}_{Total} = \alpha \mathcal{L}_{supcon} + \beta\mathcal{L}_{entropy} + \gamma\mathcal{L}_{sparsity} 
\end{equation}
\noindent where $\alpha$, $\beta$ and $\gamma$ are weights to control the impact of each loss term.
Finally, after training the model using the combined loss $\mathcal{L}_{\text{Total}}$, we collect the sparse representation vectors for all images in the training set. These vectors are binarized—each value thresholded at $0.5$—to form a binary table, where each row corresponds to an image and each column to a neuron in the final linear layer. This binarization table is used as input to the FOLD-SE-M algorithm, which generates a symbolic rule-set in the form of a stratified Answer Set Program. Each predicate in the resulting rule-set corresponds directly to a neuron in the final layer, enabling symbolic reasoning over the learned representations.

\medskip\noindent\textbf{Semantic Labelling Algorithm: } To assign human-understandable meanings to these predicates, we employ the semantic labeling algorithm introduced by \cite{nesyfold}. For each neuron, we identify the top-$10$ images that activate it most strongly. We then compute attention-based heatmaps for these images to localize the spatial regions that the neuron focuses on. These heatmaps are intersected with the corresponding (pre-annotated) semantic segmentation masks—where image regions are labeled with semantic object categories. For each neuron, we compute the average Intersection-over-Union (IoU) between its heatmaps and each available concept label. The label with the highest average IoU is then assigned to the predicate corresponding to that neuron, thus grounding the symbolic rule-set in interpretable visual concepts.

After the rule-set is generated, new images can be classified by passing them through the modified ViT, which outputs a $D$-dimensional vector. This vector is then binarized by thresholding each value at $0.5$. The resulting binary vector serves as input to the FOLD-SE-M toolkit’s rule interpreter, which evaluates the symbolic rule-set in a top-down manner. For each rule, the truth values of the predicates in the body are determined based on the corresponding neuron activations in the binarized vector. The first rule that fires determines the predicted class of the image. Moreover, the s(CASP) (\cite{scasp}) goal-directed ASP interpreter can be used to obtain a justification of the prediction.

\section{Experiments}
We conducted experiments to address the following research questions:

\medskip\noindent\textbf{Q1:} How does the classification accuracy of the neuro-symbolic model produced by the NeSyViT framework compare to that of a standard vanilla Vision Transformer (ViT)?

\medskip\noindent\textbf{Q2:} What is the typical size of the rule-set generated by the NeSyViT framework, and how compact are the resulting rules?

\medskip\noindent\textbf{Q3:} How does our framework scale as the number of classes increases?

\medskip\noindent\textbf{Q4:} What do the neuron activation patterns look like for each class, and how well do they align with human-interpretable semantic concepts?

\medskip\noindent\textbf{Q5:} How well can the semantic labelling algorithm meant for CNN-based frameworks such as NeSyFOLD, be directly adopted for our NeSyViT framework?

\medskip\noindent\textbf{[Q1, Q2, Q3] Classification Accuracy, Rule-set Size and Scalability:}
The central goal of the NeSyViT framework is to produce a neuro-symbolic model that maintains high classification accuracy while offering symbolic interpretability through a compact rule-set. Ideally, the accuracy of the interpretable model should match or exceed that of its purely neural counterpart (i.e., the vanilla ViT), and the resulting rule-set should remain as concise as possible. We use rule-set size as a proxy for interpretability, based on the findings of Lage et al.~(\cite{rulesetinterpretability}), who demonstrated through human-subject evaluations that larger rule-sets are significantly more difficult to interpret. Thus, achieving high accuracy alongside a small rule-set is critical for balancing performance and human-understandability.

\medskip\noindent\textbf{Setup: } We use the ViT-Base architecture from the \texttt{timm} library, which processes input images of size $224 \times 224$ using non-overlapping $16 \times 16$ patches. The model consists of 12 Transformer blocks, each with 12 attention heads, and produces a \texttt{[CLS]} token embedding of dimension 768. As described earlier, we modify this architecture by removing the final classification head and replacing it with a single linear layer of output dimension 128. All experiments are conducted using this configuration with weights pretrained on ImageNet-1k to construct the NeSy model. The vanilla ViT refers to the same base architecture without any modifications.
We used the AdamW optimizer along with a cosine annealing learning rate scheduler. The model was trained for $50$ epochs using the combined loss consisting of supervised contrastive loss, entropy minimization, and L1 sparsity, as previously described. For rule-set generation, we employed the FOLD-SE-M algorithm. The weights assigned to each loss component, as well as additional training and architectural hyperparameters, are provided in the Appendix.

We evaluated both our NeSy model and the vanilla ViT on subsets of two benchmark datasets (\ref{tb_main}): the \emph{Places} dataset~(\cite{zhou2017places}), which contains images of various indoor and outdoor scenes, and the \emph{German Traffic Sign Recognition Benchmark} (GTSRB)~(\cite{gtsrb}), which consists of images of traffic signposts. From the \emph{Places} dataset, we constructed multiple class subsets of increasing number of classes to gauge the scalability of NeSyViT: two classes (\textit{P2}), three classes (\textit{P3.1}), five classes (\textit{P5}), and ten classes (\textit{P10}). The \textit{P2} subset includes \textit{bathroom} and \textit{bedroom} images. \textit{P3.1} extends this by adding \textit{kitchen}, while \textit{P5} further incorporates \textit{dining room} and \textit{living room}. \textit{P10} adds five additional classes: \textit{home office}, \textit{office}, \textit{waiting room}, \textit{conference room}, and \textit{hotel room}. Additionally, we include two alternative three-class subsets: \textit{P3.2} with \textit{desert road}, \textit{forest road}, and \textit{street}, and \textit{P3.3} with \textit{desert road}, \textit{driveway}, and \textit{highway}. For each class, we sampled $5,000$ images, creating a $4$k/$1$k train-test split, and used the official validation set without modification.

The \textit{GTSRB} (\textit{GT43}) dataset contains $43$ traffic sign classes. We used the official test set of $12.6k$ images and performed an $80/20$ train-validation split on the remaining data, resulting in approximately $21k$ training and $5k$ validation images.

We run each experiment $5$ times with random train-test splits and then report the average metrics in Table \ref{tb_main}. We closely follow the experimental setup proposed by Padalkar et al.(\cite{nesyfold}) for evaluating the NeSyFOLD framework, which uses FOLD-SE-M to extract rule-sets from CNNs. Specifically, we compare the drop (or gain) in accuracy observed while using NeSyFOLD as compared to the vanilla CNN, against the corresponding change in accuracy when using NeSyViT as compared to the Vanilla ViT. This relative comparison highlights the effectiveness of our approach and is summarized in Table \ref{tb_relative_acc}.

\begin{table}[t]
\fontsize{9}{10}\selectfont
\begin{tabular}{@{}rlllll@{}}
\toprule
\multicolumn{1}{l}{Data} & Model & Accuracy (\%) & Rules & Unique Predicates & Size \\ \midrule
\multirow{2}{*}{\textit{P2}} & NeSyViT & $\textbf{100} \pm \textbf{0.00}$ & $2 \pm 0.00$ & $2 \pm 0.00$ & $2 \pm 0.00$ \\
& Vanilla & $99 \pm 0.00$ & - & - & - \\ \cmidrule(lr){1-6}
\multirow{2}{*}{\textit{P3.1}} & NeSyViT & $\textbf{99} \pm \textbf{0.00}$ & $3 \pm 0.00$ & $3 \pm 0.00$ & $3 \pm 0.00$ \\
& Vanilla & $98 \pm 0.00$ & - & - & - \\ \cmidrule(lr){1-6}
\multirow{2}{*}{\textit{P3.2}} & NeSyViT & $\textbf{99} \pm \textbf{0.00}$ & $3 \pm 0.00$ & $3 \pm 0.40$ & $3 \pm 0.00$ \\
& Vanilla & $97 \pm 0.00$ & - & - & - \\ \cmidrule(lr){1-6}
\multirow{2}{*}{\textit{P3.3}} & NeSyViT & $\textbf{98} \pm \textbf{0.00}$ & $5 \pm 0.40$ & $3 \pm 0.49$ & $5 \pm 0.40$ \\
& Vanilla & $91 \pm 0.00$ & - & - & - \\ \cmidrule(lr){1-6}
\multirow{2}{*}{\textit{P5}} & NeSyViT & $\textbf{98} \pm \textbf{0.00}$ & $5 \pm 0.00$ & $5 \pm 0.40$ & $5 \pm 0.80$ \\
& Vanilla & $90 \pm 0.00$ & - & - & - \\ \cmidrule(lr){1-6}
\multirow{2}{*}{\textit{P10}} & NeSyViT & $\textbf{94} \pm \textbf{0.00}$ & $10 \pm 0.49$ & $12 \pm 1.47$ & $14 \pm 1.67$ \\
& Vanilla & $76  \pm 0.00$ & - & - & - \\ \cmidrule(lr){1-6}
\multirow{2}{*}{\textit{GT43}} & NeSyViT & $98 \pm 0.00$ & $43 \pm 0.00$ & $44 \pm 1.90$ & $51 \pm 3.12$ \\
& Vanilla & $\textbf{99}  \pm \textbf{0.00}$ & - & - & - \\ \cmidrule(lr){1-6}
\multirow{2}{*}{\textit{Mean Stats}} & NeSyViT & $\textbf{98} \pm \textbf{0.00}$ & $10.14 \pm 0.13$ & $10.30 \pm 0.67$ & $11.85 \pm 0.86$ \\
& Vanilla & $92.86 \pm 0.00$ & - & - & - \\ \cmidrule(lr){1-6}
\end{tabular}
\caption{Comparison between the NeSyViT and the Vanilla ViT. Bold values are better.}
\label{tb_main}
\end{table}

\begin{table}[t]
\centering
\fontsize{9}{10}\selectfont
\begin{tabular}{@{}lcc|cc@{}}
\toprule
\textbf{Data} & \multicolumn{2}{c|}{\textbf{\% Accuracy Change}} & \multicolumn{2}{c}{\textbf{Rule-set Size}} \\
              & NeSyFOLD & NeSyViT & NeSyFOLD & NeSyViT \\ \midrule
\textit{P2}         & $-4$     & $\textbf{+1}$    & $12$     & $\textbf{2}$      \\
\textit{P3.1}    & $-7$   & $\textbf{+1}$    & $16$     & $\textbf{3}$     \\
\textit{P3.2}         & $-4$     & $\textbf{+2}$    & $7$     & $\textbf{3}$      \\
\textit{P3.3}    & $-7$     & $\textbf{+7}$    & $23$     & $\textbf{5}$     \\
\textit{P5}         & $-15$     & $\textbf{+8}$    & $30$     & $\textbf{5}$      \\
\textit{P10}    & $-21$     & $\textbf{+18}$    & $65$     & $\textbf{14}$     \\
\textit{GT43}         & $-13$     & $\textbf{-1}$    & $99$     & $\textbf{51}$      \\\midrule
\textit{Mean Stats}    & $-10.14$     & $\textbf{+5.14}$    & $36$     & $\textbf{11.85}$     \\
% Add more rows here
\bottomrule
\end{tabular}
\caption{Comparison of relative \% accuracy change w.r.t. the vanilla model and rule-set size between NeSyFOLD and NeSyViT.}
\label{tb_relative_acc}
\end{table}

\medskip\noindent\textbf{Result: } Table \ref{tb_main} presents a comparison between the NeSy model generated using NeSyViT and the Vanilla ViT. All metrics are reported over five independent runs. Accuracy is computed on the test set for both models. The ``Rules" column reports the average and standard deviation of the number of rules generated. ``Unique Predicates" indicates the number of distinct predicates used across the rule-set, and ``Size" refers to the total number of predicate occurrences in the bodies of all rules. Recall that we use rule-set size as a proxy for interpretability—smaller rule-sets are generally easier to understand.

The most significant observation is that our method achieves higher average accuracy than the Vanilla ViT. This is particularly noteworthy, as extracting interpretable rule-sets by binarizing internal representations typically results in some loss of information and a subsequent drop in performance. However, our approach mitigates this issue through the use of an entropy loss, which encourages the outputs of the sparse concept layer to be close to either $0$ or $1$ during training. As a result, thresholding these outputs at $0.5$ post-training introduces minimal distortion. Furthermore, the supervised contrastive loss ensures that the resulting binary vectors are well-clustered by class, enabling FOLD-SE-M to learn accurate and class-discriminative rule-sets. Finally, the L1 sparsity loss encourages only a small subset of neurons to activate for each image, resulting in compact and interpretable rule-sets. Notice that for the \textit{P10} dataset, the accuracy improvement over vanilla is $\textbf{18\%}$, which is a huge improvement and shows the merit of our approach when scaling to larger number of classes.

To provide context for the significance of our results, Table \ref{tb_relative_acc} presents the percentage change in accuracy of the NeSy model relative to its vanilla counterpart for both NeSyViT and NeSyFOLD (which uses a CNN instead of a ViT). We also report the average rule-set size generated by each framework. Notably, NeSyViT improves upon the accuracy of the Vanilla ViT by an average of $\textbf{5.14\%}$—a substantial gain, especially when contrasted with NeSyFOLD, which suffers an average accuracy drop of $\textbf{10.14\%}$ compared to the vanilla CNN. Additionally, NeSyFOLD shows a consistent degradation in performance as the number of classes increases, largely due to the information loss introduced by post-hoc binarization. In contrast, our method consistently yields accuracy improvements across datasets, with the sole exception of \textit{GT43}, where we observe a minor drop of 1 percentage point.

Furthermore, the rule-sets generated by NeSyViT are significantly more compact, with an average size of $11.85$, which is $\textbf{67\%}$ smaller than those produced by NeSyFOLD—further emphasizing the interpretability of our approach.

\begin{figure}[t]
    \centering
    \includegraphics[width=\linewidth, trim=0 10 250 50, clip]{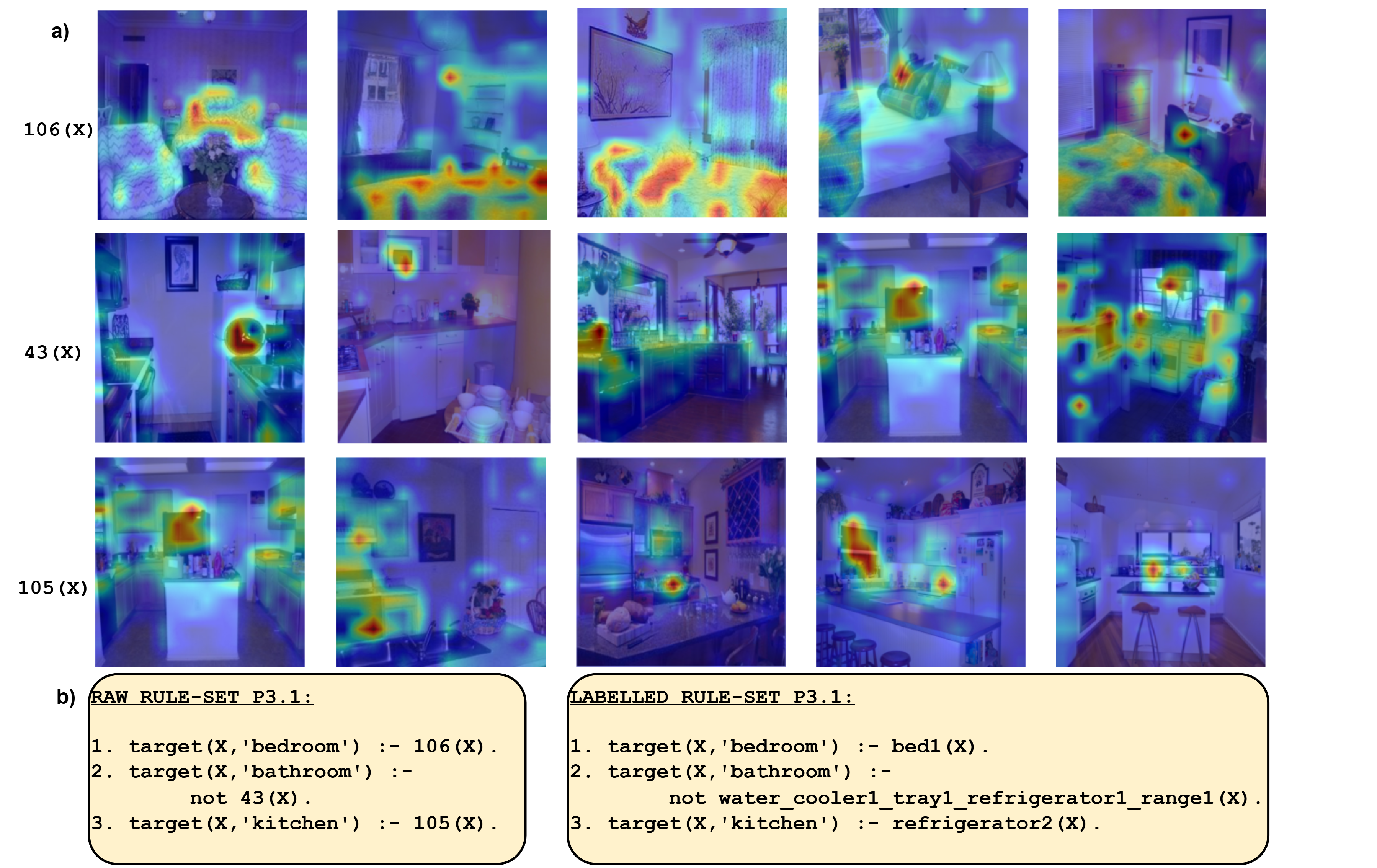}
    \caption{\textbf{a)} The top-5 images overlayed with activation heat-maps for neurons \texttt{106, 43} and \texttt{105} when rules are extracted for the \textit{P3.1} dataset containing classes ``bathroom", ``bedroom" and ``kitchen". \textbf{b)} The raw rule-set and the labelled rule-set when NeSyViT is employed on the \textit{P3.1} dataset.}
    \label{fig_activation_maps}
\end{figure}

\medskip\noindent\textbf{[Q4, Q5] Neuron Activation Patterns and Automatic Semantic Labelling Efficacy: } A key component of interpretability in neuro-symbolic frameworks is the alignment of individual neurons with semantically meaningful concepts relevant to specific classes.

\medskip\noindent\textbf{Setup: } To evaluate this alignment, we apply the previously described, semantic labelling algorithm to assign human-interpretable concept names to each predicate in the rule-set, where each predicate corresponds to a neuron in the final sparse concept layer. Initially, the raw rule-set produced by the FOLD-SE-M algorithm uses only neuron indices as predicate names. The semantic labelling process replaces these with descriptive concept labels derived by analyzing the top-10 most activating images per neuron. These images are overlaid with attention-based heatmaps and compared with pre-annotated semantic segmentation masks to identify the dominant concept for each neuron.

In Figure~\ref{fig_activation_maps}a), we visualize the top-$5$ most activating images for each predicate in the labelled rule-set for the \textit{P3.1} dataset, which contains the classes ``bathroom", ``bedroom", and ``kitchen." Figure~\ref{fig_activation_maps}b) shows a side-by-side comparison of the raw and semantically labelled rule-sets, illustrating how the abstract neuron indices are transformed into interpretable concept-based rules.

\medskip\noindent\textbf{Result: } Examining the labelled rule-set in Figure~\ref{fig_activation_maps}b), we observe that just three rules are sufficient to achieve a classification accuracy of $99\%$. Consider Rule 2: \texttt{target(X, `bathroom') :- not water\_cooler1\_tray1\_refrigerator1\_range1(X)}. At first glance, this rule may appear counterintuitive—why would the absence of several kitchen-related concepts imply that the image depicts a bathroom rather than a bedroom? The explanation lies in the execution strategy of FOLD-SE-M, which evaluates rules in a top-down fashion. Rule 2 only activates if Rule 1 fails to fire. Rule 1 is: \texttt{target(X, `bedroom') :- bed1(X)}. Thus, if the image lacks the ``bed" concept, Rule 1 is bypassed, and Rule 2 checks for the absence of kitchen concepts; if none are present, the image is classified as a ``bathroom." Rule 3: \texttt{target(X, `kitchen') :- refrigerator2(X)} fires only if the previous two rules do not apply and the ``refrigerator" concept is present, thereby classifying the image as kitchen."

While the logical flow of the rules is valid, the quality of semantic labelling is less reliable. The labelling pipeline we adapted from NeSyFOLD was designed for CNNs, where modular filters make concept isolation more straightforward. In ViTs, however, neurons often attend to multiple regions and concepts simultaneously, making the learned representations less disentangled. As a result, some neuron labels can be misleading. This limitation is evident in Figure~\ref{fig_activation_maps}a): neuron $106$ consistently activates for the ``bed" concept in bedroom images, but neurons $43$ and $105$ do not display clear concept selectivity and instead respond to mixed features. This highlights the challenge of isolating concepts in self-attention-based architectures.

While the automatic labelling approach may have limitations, the visualization of neuron activation heatmaps remains a powerful tool. These heatmaps allow for manual inspection of concept associations and provide valuable interpretability cues. We believe this combination of automated rule generation and visual validation offers a promising direction. Also, since the extracted rule-set is an Answer Set Program we can get justification of any prediction using the s(CASP) (\cite{scasp}) ASP interpreter. For completeness, we provide the labelled rule-sets for all datasets used in our experiments in Table~\ref{tb_main}, in the Appendix.

\section{Related Works}
The integration of symbolic reasoning with deep learning has led to various approaches for extracting interpretable rules from neural networks. While initial efforts focused on convolutional neural networks (CNNs), recent studies have begun exploring Vision Transformers (ViTs).
VisionLogic (\cite{visionlogic}) introduces a framework that transforms neurons in the final fully connected layer of deep vision models into predicates, grounding them into visual concepts through causal validation. ViT-NeT (\cite{vitnet}) presents a neural tree decoder that interprets the decision-making process of ViTs. By routing images hierarchically through a tree structure, it provides transparent and interpretable classifications, addressing the trade-off between model complexity and interoperability. These methods primarily focus on post-hoc interpretability or rely on additional structures for explanation. In contrast, our approach integrates a sparse linear layer directly into the ViT architecture, enabling the extraction of executable logic programs without auxiliary components.

Enhancing the interpretability of ViTs has been a subject of extensive research, leading to various methodologies. INTR (\cite{intr}) proposes a proactive approach, asking each class to search for itself in an image. This idea is realized via a Transformer encoder-decoder where “class-specific” queries (one for each class) are learnt as input to the decoder, enabling each class to localize its patterns in an image via cross-attention. Each class is attended to very distinctly hence the cross-attention weights provide a interpretation of the prediction.
% ProtoViT (\cite{protovit}) integrates deep learning with case-based reasoning by classifying images through comparisons with learned prototypes. This method provides explanations in the form of ``this looks like that," offering intuitive insights into the model's classifications. 
LeGrad (\cite{legrad}) introduces an explainability method specifically designed for ViTs. By computing gradients with respect to attention maps and aggregating signals across layers, LeGrad produces explainability maps that offer insights into the model's focus areas during decision-making. While these approaches enhance the transparency of ViTs, they do not produce symbolic or executable explanations. Our method differs by generating logic programs that serve as the final decision layer, enabling symbolic reasoning directly from the model's internal representations.

The pursuit of sparse and disentangled representations in ViTs has led to innovative methodologies. Recent work integrates sparse autoencoders (SAEs) with Vision Transformers (ViTs) to improve interpretability. \cite{saeclip} trained SAEs on CLIP's ViT and found that manipulating a small set of steerable features can enhance performance and robustness. PatchSAE (\cite{saevit}) introduces method to extract granular visual concepts and study how these influence predictions, showing that most adaptation gains stem from existing features in the pre-trained model.
Our approach builds upon these concepts by integrating a sparse linear layer into the ViT architecture and using it for extracting a symbolic rule-set which gives a global explanation of the model and a justification can be obtained for each prediction.

\section{Conclusion and Future Work}
In this work, we proposed a novel neuro-symbolic framework for interpretable image classification using Vision Transformers (ViTs). By replacing the final classification layer with a linear layer (sparse concept layer) producing sigmoid outputs, and introducing a unified loss function—comprising supervised contrastive loss for better class separation in latent space, entropy loss for sharper binarization, and sparsity loss for concept selectivity—we generated sparse binary vectors that represent images in a disentangled manner. These vectors serve as inputs to the FOLD-SE-M algorithm, which produces stratified Answer Set Programs in the form of concise rule-sets. The resulting neuro-symbolic (NeSy) model, composed of the modified ViT and the learned rule-set, outperforms the vanilla ViT by an average of \textbf{5.14\%} in accuracy.

We compared our framework, \textbf{NeSyViT}, with \textbf{NeSyFOLD}, a similar approach for CNNs, and found that NeSyViT produces rule-sets that are \textbf{67\% smaller} on average while improving upon the baseline ViT's accuracy—whereas NeSyFOLD suffers a \textbf{10.14\%} drop in performance compared to its CNN counterpart. These results demonstrate the potential of attention-based architectures for interpretable, logic-driven classification.

Finally, we investigated the semantic interpretability of neurons in our framework. We observed that the automatic semantic labelling algorithm used in NeSyFOLD—based on overlap with segmentation masks—struggles in ViTs due to limited neuron monosemanticity. However, we argue that this challenge can be partially mitigated by visualizing attention-guided heatmaps for the small set of neurons that appear in the rule-set, enabling manual inspection and insight into the learned concepts.

Future work will focus on improving neuron disentanglement and enhancing monosemanticity using architectural or training refinements. Another promising direction explored in CNN-based interpretability methods is bias correction using extracted rule-sets \cite{nesybicor} which could be the natural next step for this work. We also plan to explore using multimodal LLMs like GPT-4o for concept labeling, enabling an automatic semantic annotation pipeline that does not rely on pixel-level segmentation masks.

\bibliographystyle{tlplike}
\bibliography{main}

\newpage
\appendix
\centerline{\huge Appendix}

\noindent Our paper above shows the labeled rule-set for only one dataset i.e. \textit{P3.1} so here we show the labelled rule-sets for all the other datasets as well. 
% We also show the attention heat-maps for the top-5 images for each predicate in Run $1$ of the \textit{P10} dataset.
Also we show the hyperparameters used for training and the rule generation using FOLD-SE-M.

% \noindent Note that the code-base for our work can be found at:\\
% https://anonymous.4open.science/r/ICLP2025\_code-3F44/ICLP\_2025\_code/src/README.md \\(please check the underscores when pasting in a browser)

\section{Labelled rule-sets}
% We ran PLACES2 ($ratio = 0.3$, $tail = 5e^{-3}, margin = 0.05$), PLACES5 ($ratio = 0.8$, $tail = 5e^{-3}, margin = 0.05$), ``dedrh" ($ratio = 0.8$, $tail = 5e^{-3}, margin = 0.1$), ``dedrs" ($ratio = 0.8$, $tail = 5e^{-3}, margin = 0.1$) and ``defs" ($ratio = 0.8$, $tail = 5e^{-3}, margin = 0.1$).

\medskip\noindent\textbf{P2 \{``bathroom", ``bedroom"\}:}\\
\small{
\begin{verbatim}
target(X,'bathroom') :- bathroom_tiles1_shower_screen1(X).
target(X,'bedroom') :- bed1_duvet1(X).

\end{verbatim}
}

\medskip\noindent\textbf{P3.1 \{``bathroom", ``bedroom", ``kitchen"\}:}\\
\small{
\begin{verbatim}
target(X,'bedroom') :- not shower_screen1(X).
target(X,'bathroom') :- 
    not refrigerator1_kitchen_island1_range1_countertop1_person1_wall1_air_conditioning1(X).
target(X,'kitchen') :- not bathtub1_mirror1(X).
\end{verbatim}
}

\medskip\noindent\textbf{P3.2 \{``desert road", ``forest road", ``street"\}:}
\small{
\begin{verbatim}
target(X,'desert_road') :- field1_grass1_desert1(X).
target(X,'street') :- food_cart1(X).
target(X,'forest_road') :- not building1(X).


\end{verbatim}

\medskip\noindent\textbf{P3.3 \{``desert road", ``driveway", ``highway"\}:}
\small{
\begin{verbatim}
target(X,'highway') :- not buildings1(X).
target(X,'driveway') :- not hill1_ground3(X).
target(X,'desert_road') :- field1_ground1(X).
target(X,'highway') :- buildings1(X).
target(X,'desert_road') :- ground2(X).

\end{verbatim}
}

\medskip\noindent\textbf{P5 \{``bathroom", ``bedroom", ``kitchen", ``dining room", ``living room"\}:}\\

\small{
\begin{verbatim}
target(X,'bathroom') :- toilet1(X).
target(X,'living_room') :- fireplace1(X).
target(X,'dining_room') :- not bed2_railing1(X).
target(X,'bedroom') :- bed1(X).
target(X,'kitchen') :- top1(X).

\end{verbatim}
}

\medskip\noindent\textbf{P10 \{``bathroom", ``bedroom", ``kitchen", ``dining room", ``living room", ``home office", ``office", ``waiting room", ``conference room", ``hotel room"\}:}\\
\small{
\begin{verbatim}
target(X,'bedroom') :- bed1(X), not bed2(X).
target(X,'bathroom') :- shower_screen1_mirror1(X).
target(X,'dining_room') :- tablecloth1_bar1(X), not work_surface1_front1_trash_can1(X).
target(X,'office') :- tablecloth1_bar1(X).
target(X,'living_room') :- shelves1(X), not table1_lectern1_floor2_top1(X).
target(X,'conference_room') :- shelves1(X).
target(X,'kitchen') :- floor1(X).
target(X,'hotel_room') :- sofa1_seat1_armchair1_door1_wall1(X).
target(X,'home_office') :- bed1(X).
target(X,'waiting_room') :- bed2(X), not bathtub1(X).

\end{verbatim}
}

\section{Neuron Attention-Maps for \textit{P10}}
\small{
\begin{verbatim}
label(X,'bedroom') :- 14(X), not 83(X).
label(X,'bathroom') :- 81(X).
label(X,'dining_room') :- 122(X), not 97(X).
label(X,'office') :- 122(X).
label(X,'living_room') :- 43(X), not 45(X).
label(X,'conference_room') :- 43(X).
label(X,'kitchen') :- 41(X).
label(X,'hotel_room') :- 50(X).
label(X,'home_office') :- 14(X).
label(X,'waiting_room') :- 83(X), not 5(X).
\end{verbatim}
}
% \begin{figure}
%     \centering
%     \includegraphics[width=\linewidth,trim={0 0 100cm 0},clip]{Images/P10_activations.png}
%     \caption{The neuron activations for all the predicates in the rule-set for \textit{P10}}
%     \label{fig_P10activations}
% \end{figure}
\section{Hyperparameters}
\noindent\textbf{NeSyViT: }
For the subsets of the Places dataset i.e. \textit{P2}, \textit{P3.1}, \textit{P3.2}, \textit{P3.3}, \textit{P5}, \textit{P10}
We used a learning rate of $5 \times 10^{-6}$ and l2 weight decay of $5e^{-3}$ with the AdamW optimizer, batch size of $32$, patience of $10$ epochs for the LR-Scheduler, LR decay factor of $0.5$.

The weights for the supcon loss, sparsity loss and entropy loss were $2, 1$ and $1$ respectively.

For the \textit{GT43} dataset we changed the learning rate to $5 \times 10^{-5}$ and the weights for the supcon loss, sparsity loss and entropy loss were $4, 0.001$ and $0.001$ respectively. Every other hyperparameter was the same.
 
For all datasets we used a ratio of $0.8$ and a tail of $5e^{-3}$ for the FOLD-SE-M algorithm.

\noindent\textbf{Vanilla ViT: } For all datasets we used a learning rate of $5 \times 10^{-6}$, batch size of $32$, patience of $10$ epochs for the LR-Scheduler, LR decay factor of $0.5$ and l2 weight decay of $5e^{-3}$. The loss used was standard crossentropy loss.

\end{document}